\pdfoutput=1

\documentclass[11pt]{article}

\usepackage[final]{acl}
\usepackage{booktabs}
\usepackage{times}
\usepackage{latexsym}
\usepackage{enumitem} 
\usepackage[T1]{fontenc}
\usepackage[utf8]{inputenc}
\usepackage{microtype}
\usepackage{inconsolata}
\usepackage{float}

\usepackage{graphicx}
\usepackage{caption} 
\usepackage{colortbl}
\usepackage{lipsum}
\usepackage{times}
\usepackage{latexsym}
\usepackage{subcaption}  
\usepackage[T1]{fontenc}
\usepackage{amsmath}
\usepackage[utf8]{inputenc}
\usepackage{enumitem}
\usepackage{microtype}
\usepackage{xcolor}
\usepackage[most]{tcolorbox}
\usepackage{float} 

\usepackage{hyperref}

\usepackage{multirow}
\usepackage[T1]{fontenc}
\usepackage[normalem]{ulem}
\useunder{\uline}{\ul}{}
\usepackage{graphicx}
\usepackage{longtable}
\usepackage{amsfonts}
\usepackage{amssymb}
\usepackage{pifont}
\usepackage{array} 
\usepackage[linesnumbered,ruled,vlined]{algorithm2e} 
\usepackage{mdframed}

\usepackage{tcolorbox}
\usepackage{xcolor}
\usepackage[table]{xcolor} 
\usepackage{makecell}

\definecolor{lightpink}{HTML}{FFF2F2}

\definecolor{deepblue}{RGB}{0, 0, 139}
\definecolor{deepgreen}{RGB}{0, 100, 0}   
\definecolor{deepred}{RGB}{139, 0, 0}     
\definecolor{deeppurple}{RGB}{75, 0, 130}  

\definecolor{linkblue}{HTML}{2221FC}
\definecolor{lightyellow}{HTML}{FEFDE9}
\definecolor{lightper}{HTML}{F2F2FF}
\definecolor{sum}{HTML}{FFAF59}
\definecolor{arg}{HTML}{4497B8}
\definecolor{eval}{HTML}{778BD2}
\definecolor{red}{HTML}{FF0000}
\definecolor{green}{HTML}{689157}
\definecolor{down}{rgb}{0.8, 0.0, 0.0}
\definecolor{up}{HTML}{4E9C71}
\definecolor{tabletitle}{HTML}{DAD9E4}
\definecolor{linkgreen}{HTML}{3C7E7F}

\usepackage{xurl}

\def\method{\texttt{EasyRL}}

\title{Easy Samples Are All You Need: Self-Evolving LLMs via Data-Efficient Reinforcement Learning}

\author{
Zhiyin Yu\textsuperscript{ \rm $\heartsuit$ \ding{171}}, 
Bo Zhang\textsuperscript{\ding{171} \textdagger}, 
Qibin Hou\textsuperscript{\rm $\diamondsuit$ }, 
Zhonghai Wu\textsuperscript{\rm $\heartsuit$ \textdagger}, 
Xiao Luo\textsuperscript{\ding{168}},
Lei Bai\textsuperscript{\ding{171}}
\\
{\textsuperscript{\rm $\heartsuit$}{Peking University}} 
{\textsuperscript{\ding{171}}Shanghai Artificial Intelligence Laboratory} \\
{\textsuperscript{\rm $\diamondsuit$}Nankai University}
{\textsuperscript{\ding{168}}University of Wisconsin–Madison}  \\
\small\texttt{zhiyinyu25@stu.pku.edu.cn, zhangbo@pjlab.org.cn, houqb@nankai.edu.cn}\\
\small\texttt{wuzh@pku.edu.cn, xiao.luo@wisc.edu, bailei@pjlab.org.cn}\\
{\tt Github Repository: \url{https://github.com/YuZhiyin/EasyRL}.}
}

\makeatletter
\def\blfootnote{\xdef\@thefnmark{}\@footnotetext}
\makeatother

\begin{document}
\maketitle
\begin{abstract}
Previous LLMs-based RL studies typically follow either supervised learning with high annotation costs, or unsupervised paradigms using voting or entropy-based rewards. However, their performance remains far from satisfactory due to the substantial annotation cost and issues such as model collapse or reward hacking. To address these issues, we introduce a new perspective inspired by cognitive learning theory and propose a novel approach called \method{}. The core of \method{} is to simulate the human cognitive acquisition curve by integrating reliable knowledge transfer from easy labeled data with a progressive divide-and-conquer strategy that tackles increasingly difficult unlabeled data. Specifically, we initialize a warm-up model using supervised RL with few-shot labeled data. This is followed by a divide-and-conquer pseudo-labeling strategy on difficult unlabeled data, combining consistency-based selection for low-uncertainty cases and reflection-based resolution for medium-uncertainty cases. Finally, difficulty-progressive self-training with iterative pseudo-labeling and RL further strengthens the model’s reasoning capability. \method{} provides a unified self-evolving framework that facilitates data-efficient post-training of LLMs. Experimental results on mathematical and scientific benchmarks demonstrate that \method{}, using only 10\% of easy labeled data, consistently outperforms state-of-the-art baselines.

\end{abstract}
\blfootnote{ \textsuperscript{\textdagger} Corresponding authors. }

\section{Introduction}

Large language models (LLMs) have demonstrated remarkable capabilities across diverse domains such as mathematical reasoning~\citep{cui2025processreinforcementimplicitrewards,deepscaler2025,guan2025rstarmath}, scientific research~\citep{internagentteam2025internagentagentscientist,yu-etal-2025-scrag}, and code generation~\citep{guo2024deepseekcoderlargelanguagemodel,qian-etal-2024-chatdev}. Recent advances including DeepSeek-R1~\citep{deepseekai2025deepseekr1incentivizingreasoningcapability}, OpenAI's o1~\citep{openai2024openaio1card}, and Kimi-1.5~\citep{kimiteam2025kimik15scalingreinforcement} have shown that Reinforcement Learning (RL), as a promising post-training paradigm, can effectively elicit and enhance the reasoning abilities of LLMs. Intriguingly, LLMs trained with RL exhibit emergent cognitive behaviors such as self-reflection~\citep{gandhi2025cognitivebehaviorsenableselfimproving}, and spontaneous “aha moments,” while also achieving strong generalization across diverse downstream tasks~\citep{lambert2025tulu3pushingfrontiers}.

In recent advances of LLM reinforcement learning for reasoning, existing approaches can be broadly categorized into supervised and unsupervised paradigms~\citep{zhang2025freelunchrethinkinginternal}. Supervised methods rely on outcome-based rewards, derived either from verifiable answers or pre-trained reward models~\citep{lyu2025exploringlimitoutcomereward,ouyang2022traininglanguagemodelsfollow}. While effective, these methods heavily depend on human-labeled data or reward model supervision, resulting in substantial annotation cost~\citep{zhang2025consistentpathsleadtruth} and limited scalability~\citep{yuan2024selfrewarding,liu2025surveydirectpreferenceoptimization}. On the other hand, unsupervised methods attempt to construct reward signals through voting~\citep{zuo2025ttrltesttimereinforcementlearning} or entropy estimation~\citep{zhang2025right}. However, their performance gains are often marginal and susceptible to issues such as model collapse~\citep{shumailov2024ai} or reward hacking~\citep{shafayat2025largereasoningmodelsselftrain}, and they struggle to generalize across diverse models~\citep{shao2025spuriousrewardsrethinkingtraining}, making stable self-evolution difficult to achieve.

Inspired by Mind in Society~\citep{ef4d7fb0-848f-3480-8634-d49a5f5c57df}, cognitive development and skill acquisition are known to follow a from-easy-to-hard trajectory. Vygotsky’s seminal Zone of Proximal Development (ZPD) theory posits learners master complex tasks by first internalizing knowledge from simple and achievable cases, and then gradually extending this knowledge to more difficult challenges with minimal external guidance. Subsequent studies further validate this principle: humans require only a small set of easy, labeled examples to induce generalizable rules~\citep{lake2016buildingmachineslearnthink}, and can transfer this foundational knowledge to tackle novel, harder problems through analogy and self-reflection~\citep{10.1162/opmi_a_00123}. This insight highlights a critical opportunity: if LLM training can be aligned with human cognitive patterns such that limited easy-labeled cases guide the model’s progression toward increasingly difficult tasks, we may achieve a data-efficient RL paradigm that balances annotation cost and generalization. Therefore, we aim to address the following question:

\begin{quote}
\vspace{-1mm}
\textit{Can LLMs gradually evolve with limited easy labeled data and abundant difficult unlabeled data?}
\vspace{-1mm}
\end{quote}

\begin{figure}[t]
    \centering
    \includegraphics[width=\columnwidth]{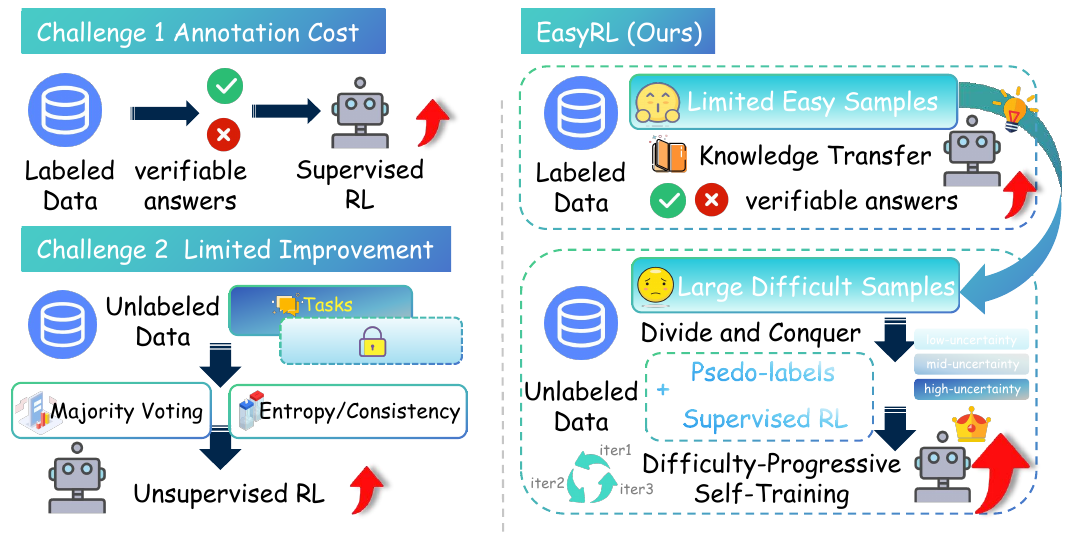}
    \caption{Comparison between (a) existing supervised and unsupervised RL approaches and (b) \method{}.}
    \vspace{-6mm}
    \label{fig:introduction}
\end{figure}

To address this question, we propose \method{}, a novel reinforcement learning framework that enables LLMs to self-evolve from limited easy cases toward more difficult reasoning tasks. Different from existing supervised and unsupervised approaches, the core of \method{} is to not only transfer reliable knowledge from easy labeled data, but also leverage difficult unlabeled data in a structured, progressively refined manner to enhance reasoning capability (see Figure~\ref{fig:introduction}). Specifically, we first transfer knowledge from labeled data to initialize a warm-up model using supervised reinforcement learning. Next, we adopt a Divide-and-Conquer strategy to construct high-quality pseudo-labeled datasets, where consistency-based selection handles low-uncertainty cases and reflection-based resolution addresses medium-uncertainty ones. To further promote self-improvement, we incorporate difficulty-progressive self-training: the model is iteratively trained on increasingly challenging pseudo-labeled samples, with selection criteria dynamically updated to follow an easy-to-hard progression aligned with human cognitive learning. Comprehensive experiments across multiple reasoning benchmarks demonstrate that \method{} significantly outperforms competitive baselines, and enables a self-evolving learning trajectory in LLMs. The main contributions are summarized as follows:

\begin{itemize}[left=0pt]
\vspace{-2mm}
\item[\ding{182}] \textit{New Perspective.} We introduce a cognitive-inspired framework that enables large language models to self-evolve from limited easy cases to more difficult reasoning tasks.
\vspace{-2mm}
\item[\ding{183}] \textit{Novel Methodology.} \method{} transfers knowledge from easy samples and progressively incorporates unlabeled data via divide-and-conquer, achieving difficulty-progressive self-training to enhance reasoning.

\vspace{-3mm}
\item[\ding{184}] \textit{Extensive Experiments.} 
We conduct extensive experiments across multiple  benchmarks, showing that \method{} is \textbf{(1) data-efficient}, using only 10\% easy labeled data while surpassing GRPO trained on the full dataset; \textbf{(2) self-evolving}, with steadily improving pseudo-label quality and an increasing focus on more difficult samples; and \textbf{(3) robust-performing}, outperforming supervised and unsupervised RL baselines and transferring effectively to out-of-domain tasks.

\end{itemize}

\begin{figure*}[t]
    \centering
    \includegraphics[width=\linewidth]{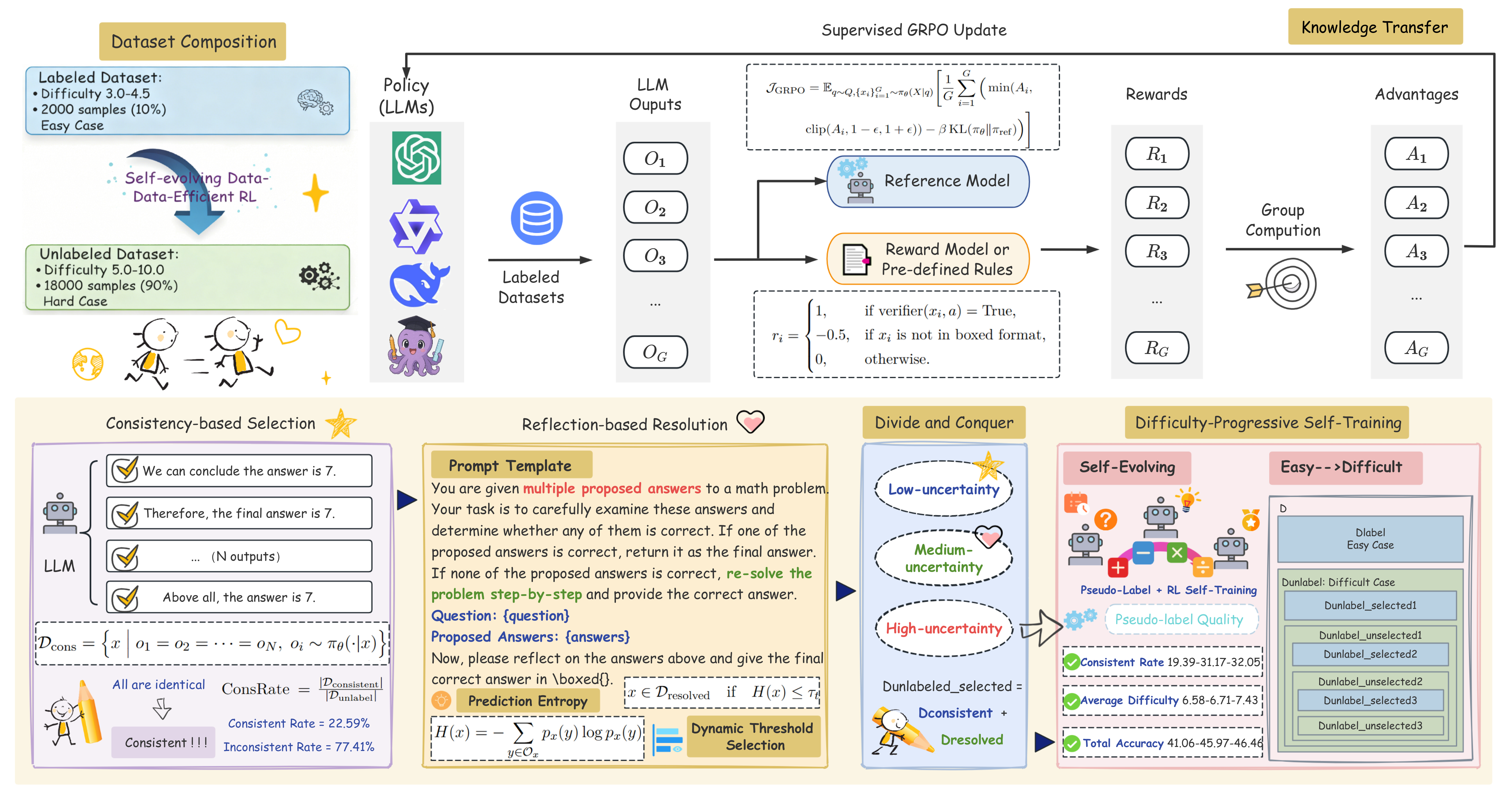}
    \vspace{-16pt}
    \caption{Overview of \method{}. The workflow consists of three stages: (1) Knowledge Transfer, which initializes a stable warm-up model from easy labeled data using supervised RL; (2) Divide and Conquer, which generates high-quality pseudo-labeled data using consistency-based selection and reflection-based resolution; and (3) Difficulty-Progressive Self-Training, which refines the model by incorporating increasingly difficult pseudo-labeled samples in a staged RL training loop.}
    \vspace{-5mm}
    \label{fig:framework}
\end{figure*}

\vspace{-3mm}

\section{Related Work}
\subsection{Reinforcement Learning for Reasoning}
Reinforcement learning with Verifiable Rewards (RLVR) has been widely utilized to strengthen the reasoning capabilities of LLMs~\citep{deepseekai2025deepseekr1incentivizingreasoningcapability,yang2025zeroguiautomatingonlinegui,yang2025ssrzerosimpleselfrewardingreinforcement,zhang2025freelunchrethinkinginternal,singh2025agenticreasoningtoolintegration,li2025torlscalingtoolintegratedrl}. Recently, an increasing body of research has focused on label-free RL, which eliminates reliance on human annotations~\citep{shao2025spuriousrewardsrethinkingtraining,zweiger2025selfadaptinglanguagemodels,xin2025surrogatesignalsformatlength}.
~\citet{zuo2025ttrltesttimereinforcementlearning,shafayat2025largereasoningmodelsselftrain,wei2025unsupervisedposttrainingmultimodalllm} adopt majority voting to generate pseudo-labels for correctness rewards. Alternatively, \citet{zhang2025right} minimize predictive entropy within a latent semantic space, while other approaches~\citep{zhao2025learningreasonexternalrewards,prabhudesai2025maximizingconfidenceimprovesreasoning,li2025confidenceneedfewshotrl,agarwal2025unreasonableeffectivenessentropyminimization} quantify model confidence via self-certainty or token-level entropy. Beyond evaluating only the final outputs, several studies incorporate intermediate reasoning states into reward modeling~\citep{zhang2025consistentpathsleadtruth,zhou2025evolvinglanguagemodelslabels}. In parallel, research on self-evolving~\citep{zhao2025absolutezeroreinforcedselfplay,zhang2025darwingodelmachineopenended} and co-evolving~\citep{wang2025coevolvingllmcoderunit,fang2025serlselfplayreinforcementlearning} RL paradigms further demonstrates the potential of autonomous adaptation. Despite their progress, self-evolving RL frameworks from a cognitive perspective remain underexplored, and our work aims to bridge this gap.

\subsection{Data-efficient LLM Training}

In recent years, achieving data-efficient post-training of LLMs with limited data has become a key research direction~\citep{luo-etal-2025-survey}. Existing approaches primarily focus on five major themes: data selection~\citep{jeong2025llmselect,liu2024sampleefficient}, data quality enhancement~\citep{10800533,10858342}, synthetic data generation~\citep{dong2024selfboostinglargelanguagemodels,10.1007/978-981-97-5501-1_30}, data distillation and compression~\citep{10.1145/3640457.3688118,pan-etal-2024-llmlingua}, and self-evolving data ecosystems~\citep{you-etal-2024-llm,NEURIPS2023_91edff07}. Recent studies have also explored data selection in reinforcement learning~\citep{li2025limrrlscaling} and even demonstrated that RL with as few as one or four samples can yield improvements~\citep{wang2025reinforcementlearningreasoninglarge,fatemi2025concisereasoningreinforcementlearning}. In contrast, our work argues that by leveraging a small set of easy samples, the model can self-evolve to tackle increasingly difficult examples, aligning with the intuition of the human learning process.

\section{Methodology}

\subsection{Problem Analysis}
In real-world scenarios, obtaining annotations for difficult reasoning tasks is often expensive and time-consuming, while collecting answers for simple problems is much easier. Consider given a small labeled dataset $\mathcal{D}_{\mathrm{label}}$ consisting of easy samples and a large unlabeled dataset $\mathcal{D}_{\mathrm{unlabel}}$ containing difficult samples. In this work, we aim to explore whether a LLM can self-evolve from limited easy cases toward more difficult tasks through data-efficient RL. Inspired by \citep{he2025deepmath103klargescalechallengingdecontaminated}, we adopt the difficulty definition provided by the AoPS\footnote{\url{https://artofproblemsolving.com/wiki/index.php/AoPS_Wiki:Competition_ratings}} rating scale (1–10). Specifically, our \method{} focuses on two objectives: (1) enhancing the model’s reasoning capability and generalization performance on test sets, and (2) improving the quality of pseudo-labels generated on unlabeled difficult samples, thereby enabling a self-improving learning cycle without additional human supervision.

\subsection{Framework Overview}

As illustrated in Figure~\ref{fig:framework}, our \method{} workflow follows a multi-step process that enables the model to evolve from easy to increasingly difficult tasks. First, we transfer knowledge from the labeled dataset containing easy samples to obtain a stable warm-up model that serves as a reliable policy prior for subsequent pseudo-label generation. Then, we use this warm-up model to pseudo-label unlabeled data and partition samples by uncertainty into low-, medium-, and high-uncertainty groups, where consistency-based selection handles low-uncertainty cases and reflection-based resolution refines medium-uncertainty ones. Finally, \method{} iteratively combines labeled data with the selected pseudo-labeled samples in a reinforcement learning training loop, gradually exposing the model to increasingly difficult samples until convergence.

\subsection{Knowledge Transfer for Foundational Capability}

To provide a stable model $\pi_{\text{warm}}$ for subsequent pseudo-label generation and self-evolution on unlabeled data $\mathcal{D}_{\text{unlabel}}$, we first transfer knowledge from the labeled data $\mathcal{D}_{\text{label}}$ through supervised reinforcement learning. Specifically, we employ the Group Relative Policy Optimization (GRPO)~\citep{shao2024deepseekmathpushinglimitsmathematical} algorithm, following DeepSeek-R1-Zero~\citep{deepseekai2025deepseekr1incentivizingreasoningcapability}, to align the model’s outputs with reference answers. The optimization objective is formulated as:
\begin{equation}
\begin{aligned}
  \mathcal{J}_{\rm GRPO} &= \mathbb{E}_{q\sim Q, \{x_i\}_{i=1}^G\sim \pi_{\theta}(X|q)} \Bigg[ 
  \frac{1}{G}\sum_{i=1}^G \Big( \min(A_i, \\
  &  {\rm clip}(A_i, 1-\epsilon, 1+\epsilon)) 
  - \beta \, {\rm KL}(\pi_\theta \| \pi_{\rm ref}) 
  \Big) \Bigg],
\end{aligned}
\label{eq:1}
\end{equation}
where $\{x_1,\ldots,x_G\}$ are a group of outputs sampled from the policy model $\pi_{\theta}$, and the KL term, controlled by the coefficient $\beta$, limits divergence from the reference policy $\pi_{\text{ref}}$. The normalized advantage $A_i$ for each output $x_i$ is computed as: $A_i = \frac{r_i - \mathrm{mean}(\{r_1,\cdots,r_G\})}{\mathrm{std}(\{r_1,\cdots,r_G\})}$. The reward $r_i$ is computed from labeled pairs $(q, a) \in \mathcal{D}_{\rm label}$, based on a correctness verifier that checks whether the model output matches the ground-truth response. Additionally, a format penalty is applied if the output is not in the expected boxed format:
\begin{equation}
r_i =
\begin{cases} 
1, & \text{if } \text{verifier}(x_i, a) = \text{True},\\[0.5mm]
-0.5, & \text{if } x_i \text{ is not in boxed format},\\[0.5mm]
0, & \text{otherwise}.\\[0.5mm]
\end{cases}
\label{eq:2}
\end{equation}

This supervised RL training embeds domain-specific knowledge from the labeled data into the model. The resulting $\pi_{\text{warm}}$ thus serves as a reliable model for pseudo-label generation and self-evolution on unlabeled data.

\subsection{Divide-and-Conquer for Reliable Pseudo-Labeling}
\label{sec:3.3}

To leverage unlabeled data, we adopt a divide-and-conquer strategy for pseudo-labeling. Specifically, we partition samples based on uncertainty of pseudo-labels: (1) \textit{Low-uncertainty}: samples with consistent outputs across multiple reasoning attempts form $D_{\text{consistent}}$ via consistency-based selection. (2) \textit{Medium-uncertainty}: samples with inconsistent outputs are re-evaluated through a reflection mechanism and filtered to form $D_{\text{resolved}}$ via reflection-based resolution. (3) \textit{High-uncertainty}: highly uncertain samples are deferred to subsequent rounds for Difficulty-Progressive Self-Training (Section~\ref{sec:3.4}). The final pseudo-labeled set is $D_{\text{unlabel\_selected}} = D_{\text{consistent}} \cup D_{\text{resolved}}$.

\medskip

\noindent \textbf{Consistency-based Selection.} For each unlabeled query $x \in \mathcal{D}_{\text{unlabel}}$, the model $\pi_{\text{warm}}$ performs $N$ independent inferences:
\begin{equation}
\mathcal{D}_{\text{cons}} = \Big\{ x \;\Big|\; o_1 = o_2 = \cdots = o_N, \; o_i \sim \pi_\theta(\cdot|x) \Big\},
\label{eq:3}
\end{equation}
where ${o_1, \ldots, o_N}$ denote the generated outputs. If all outputs are identical, the sample is regarded as consistent. The overall consistency rate is defined as:
$\text{ConsRate} = \frac{|\mathcal{D}_{\text{consistent}}|}{|\mathcal{D}_{\text{unlabel}}|},$ which reflects the model’s stability and confidence on unlabeled data.

\medskip

\noindent \textbf{Reflection-based Resolution.}
For samples that fail the consistency check, we employ a reflection mechanism to assess their reliability. Given a set of proposed answers $\mathcal{O}_x = \{o_1, \dots, o_N\}$, we compute empirical probability $p_x(y)$ for each distinct answer $y$, and define the prediction entropy as:
\begin{equation}
H(x) = - \sum_{y \in \mathcal{O}_x} p_x(y) \log p_x(y).
\label{eq:4}
\end{equation}
A dynamic threshold $\tau_t$ is used to determine whether a sample can be confidently resolved:
\begin{equation}
x \in \mathcal{D}_{\text{resolved}} 
\quad \text{if} \quad H(x) \le \tau_t,
\label{eq:5}
\end{equation}
where $\tau_t$ is a dynamic threshold, set by default to 0.3. The pseudo-label is assigned as $\hat{y}_x = \text{Reflection}(\mathcal{O}_x)$, which is the answer obtained after the reflection-based resolution, while samples with $H(x) > \tau_t$ are reserved for future rounds.

In summary, this divide-and-conquer strategy progressively constructs the reliable pseudo-labeled dataset $\mathcal{D}_{\text{unlabel\_selected}}$ while avoiding noisy supervision from high-uncertainty samples. By iteratively combining Consistency-based Selection and Reflection-based Resolution, our method ensures stable and high-quality pseudo-labels that effectively support the self-evolution process.

\subsection{Difficulty-Progressive Self-Training for Evolution}
\label{sec:3.4}

To further enable self-evolution on $\mathcal{D}_{\text{unlabel\_selected}}$, we adopt a difficulty-progressive self-training strategy that exposes the model to increasingly difficult or uncertain examples.

Specifically, we first combine the pseudo-labeled set $\mathcal{D}_{\rm unlabel\_selected}^{(0)}$ obtained via the divide-and-conquer strategy with the labeled dataset $\mathcal{D}_{\rm label}$ to perform supervised RL training. The reward reflects the correctness of pseudo-labels: the model receives 1 if its output matches the pseudo-label, 0 otherwise, with an additional penalty for incorrect format, resulting in the first intermediate model $\pi_1$.

We then iteratively apply pseudo-labeling, selection (see Section~\ref{sec:3.3}), and self-training to progressively refine the model. In each iteration, the current model $\pi_i$ generates pseudo-labels for the remaining unlabeled samples $\mathcal{D}_{\rm unlabel\_unselected}^{(i)}$, which correspond to the high-uncertainty samples from the previous round in the divide-and-conquer process. These pseudo-labeled samples are then combined with labeled data for supervised reinforcement learning to obtain the next model $\pi_{i+1}$. This process continues until convergence, producing a sequence of models $\pi_1, \pi_2, \dots, \pi_n$, each increasingly capable of handling more difficult or uncertain cases. Formally, at iteration $i$:
\begin{equation}
\pi_{i+1} = {\rm RL}(\mathcal{D}_{\rm label} \cup \mathcal{D}_{\rm unlabel\_selected}^{(i)}),
\label{eq:6}
\end{equation}
where $\mathcal{D}_{\rm unlabel\_selected}^{(i)}$ denotes the pseudo-labeled set selected by $\pi_i$ in the current iteration. This framework enables gradual evolution from easy to difficult instances, leveraging both labeled and pseudo-labeled data.

\begin{algorithm}[t]
\caption{\method{} framework}
\label{alg:SemiRL}
\SetKwInOut{Input}{Input}
\SetKwInOut{Output}{Output}

\Input{Labeled dataset $\mathcal{D}_{\text{label}}$, unlabeled dataset $\mathcal{D}_{\text{unlabel}}$, maximum iterations $I_{\max}$.}
\textcolor{linkgreen}{\texttt{// Step 1: Knowledge Transfer}}\;
Train a supervised RL model $\pi_{\text{warm}}$ on $\mathcal{D}_{\text{label}}$ using Eq.~\ref{eq:1} and~\ref{eq:2};

\textcolor{linkgreen}{\texttt{// Step 2: Divide and Conquer}}\;
\ForEach{$x \in \mathcal{D}_{\text{unlabel}}$}{
    Generate $N$ outputs $\{o_i\}_{i=1}^N$ using $\pi_{\text{warm}}$ by Eq.~\ref{eq:3}; Consistency-based Selection  to obtain $\mathcal{D}_{\text{consistent}}$;
    
    Reflection-based Resolution by Eq.~\ref{eq:4} and Eq.~\ref{eq:5} to obtain $\mathcal{D}_{\text{resolved}}$;

}
$\mathcal{D}_{\text{unlabel\_selected}} = \mathcal{D}_{\text{consistent}} \cup \mathcal{D}_{\text{resolved}}$;

\textcolor{linkgreen}{\texttt{// Step 3: Difficulty-Progressive Self-Training}}\;
\For{$i \gets 0$ \KwTo $I_{\max}$}{
    Train $\pi_i$ with RL on $\mathcal{D}_{\text{label}} \cup \mathcal{D}_{\text{unlabel\_selected}}^{(i)}$ to obtain $\pi_{i+1}$;
    
    Use $\pi_{i+1}$ to generate new pseudo-labels for $\mathcal{D}_{\text{unlabel\_unselected}}^{(i)}$ via Step~2;
    
    Update $\mathcal{D}_{\text{unlabel\_selected}}^{(i+1)}$;
}
\Output{Evolved model $\pi_{\text{final}}$.}
\end{algorithm}

\subsection{Summarization}
To summarize, the overall workflow of \method{} is presented in Algorithm~\ref{alg:SemiRL}. First, in the Knowledge Transfer stage, a supervised RL model $\pi_{\text{warm}}$ is trained on labeled data to provide a reliable policy prior for pseudo-label generation. Next, the Divide-and-Conquer stage leverages $\pi_{\text{warm}}$ to generate multiple candidate outputs for each unlabeled query and organizes them by uncertainty via consistency-based selection and reflection-based resolution, resulting in a high-quality pseudo-labeled dataset $\mathcal{D}_{\text{unlabel\_selected}}$. From a cognitive perspective, $\mathcal{D}_{\text{unlabel\_selected}}$ can be viewed as an analogue of the Zone of Proximal Development (ZPD), which characterizes tasks that slightly exceed an agent’s current capability yet remain learnable under appropriate guidance. Although ZPD is traditionally defined in human learning contexts involving more knowledgeable peers or instructors, we adapt this concept to the LLM RL setting by treating intrinsic feedback as implicit instructional signals. It allows the model to progressively expand its capability boundary, facilitating self-evolution through difficulty-progressive self-training.

\section{Experiment}
\subsection{Experimental Setup}

\begin{table*}[t]
\renewcommand\arraystretch{1.2}
\label{main_results}
\begin{center}
\resizebox{\textwidth}{!}{
\begin{tabular}{lcccccccccc}
\Xhline{1.2pt}
\rowcolor{tabletitle}  &     \multicolumn{6}{c} {\textbf{Mathematical Reasoning}} & \multicolumn{4}{c} {\textbf{Scientific Reasoning}} \\ 
\rowcolor{tabletitle}
\multirow{-2}{*}{\textbf{Methods}} & MATH & Minerva & Olympiad    & AIME24 & AMC23 & Avg.  & Biology &Chemistry & Physics &Avg.\\ \Xhline{1.2pt}
\multicolumn{11}{c}{\textit{Qwen2.5-Math-1.5B}} \\ \hline
 \rowcolor{gray!10}Vanilla Base &$65.0$&$18.0$&$28.3$&$6.7$&$45.0$&$32.6$&$0.0$&$2.3$&$1.4$&$1.5$\\
w/ Supervised GRPO&$66.4_{\textcolor{up}{\uparrow 1.4}}$&$27.9_{\textcolor{up}{\uparrow 9.9}}$&$30.7_{\textcolor{up}{\uparrow 2.4}}$&$3.3_{\textcolor{up}{\downarrow 3.4}}$&$50.0_{\textcolor{up}{\uparrow 5.0}}$&$35.7_{\textcolor{up}{\uparrow 3.1}}$&$11.4_{\textcolor{up}{\uparrow 11.4}}$&$10.0_{\textcolor{up}{\uparrow 7.7}}$&$4.3_{\textcolor{up}{\uparrow 2.9}}$&$7.9_{\textcolor{up}{\uparrow 6.4}}$ \\
\rowcolor{gray!10} w/ Unsupervised EMPO&$66.6_{\textcolor{up}{\uparrow 1.6}}$&$28.3_{\textcolor{up}{\uparrow 10.3}}$&$31.0_{\textcolor{up}{\uparrow 2.7}}$&$6.7_{\textcolor{up}{\uparrow 0.0}}$&$\textbf{60.0}_{\textcolor{up}{\uparrow 15.0}}$&$38.5_{\textcolor{up}{\uparrow 5.9}}$&$20.0_{\textcolor{up}{\uparrow 20.0}}$&$16.9_{\textcolor{up}{\uparrow 14.6}}$&$12.1_{\textcolor{up}{\uparrow 10.7}}$&$15.6_{\textcolor{up}{\uparrow 14.1}}$ \\ 
 w/ \method{} Iter1 &$69.8_{\textcolor{up}{\uparrow 4.8}}$&$28.7_{\textcolor{up}{\uparrow 10.7}}$&$31.0_{\textcolor{up}{\uparrow 2.7}}$&$3.3_{\textcolor{up}{\downarrow 3.4}}$&$52.5_{\textcolor{up}{\uparrow 7.5}}$&$37.1_{\textcolor{up}{\uparrow 4.5}}$&$27.1_{\textcolor{up}{\uparrow 27.1}}$&$14.6_{\textcolor{up}{\uparrow 12.3}}$&$10.7_{\textcolor{up}{\uparrow 9.3}}$&$15.6_{\textcolor{up}{\uparrow 14.1}}$\\
 \rowcolor{gray!10} w/ \method{} Iter2 &$70.0_{\textcolor{up}{\uparrow 5.0}}$&$\textbf{31.6}_{\textcolor{up}{\uparrow 13.6}}$&$\textbf{31.3}_{\textcolor{up}{\uparrow 3.0}}$&$10.0_{\textcolor{up}{\uparrow 3.3}}$&$50.0_{\textcolor{up}{\uparrow 5.0}}$&$38.6_{\textcolor{up}{\uparrow 6.0}}$&$24.3_{\textcolor{up}{\uparrow 24.3}}$&$\textbf{20.8}_{\textcolor{up}{\uparrow 18.5}}$&$11.4_{\textcolor{up}{\uparrow 10.0}}$&$17.6_{\textcolor{up}{\uparrow 16.1}}$\\
  w/ \method{} Iter3&$\textbf{71.2}_{\textcolor{up}{\uparrow 6.2}}$&$30.9_{\textcolor{up}{\uparrow 12.9}}$&$\textbf{31.3}_{\textcolor{up}{\uparrow 3.0}}$&$\textbf{13.3}_{\textcolor{up}{\uparrow 6.6}}$&$55.0_{\textcolor{up}{\uparrow 10.0}}$&$\textbf{40.3}_{\textcolor{up}{\uparrow 7.7}}$&$\textbf{35.7}_{\textcolor{up}{\uparrow 35.7}}$&$17.7_{\textcolor{up}{\uparrow 15.4}}$&$\textbf{12.9}_{\textcolor{up}{\uparrow 11.5}}$&$\textbf{19.4}_{\textcolor{up}{\uparrow 17.9}}$\\ \hline
\multicolumn{11}{c}{\textit{Qwen2.5-Math-7B}} \\ \hline
 \rowcolor{gray!10} Vanilla Base &$72.6$&$16.9$&$33.8$&$16.7$&$52.5$&$38.5$&$18.6$&$24.6$&$26.4$&$24.1$ \\
 Vanilla Instruct &$85.0_{\textcolor{up}{\uparrow 12.4}}$ &$41.5_{\textcolor{up}{\uparrow 24.6}}$&$40.4_{\textcolor{up}{\uparrow 6.6}}$&$16.7_{\textcolor{up}{\uparrow 0.0}}$&$60.0_{\textcolor{up}{\uparrow 7.5}}$&$48.7_{\textcolor{up}{\uparrow 10.2}}$&$28.6_{\textcolor{up}{\uparrow 10.0}}$&$23.8_{\textcolor{up}{\downarrow 0.8}}$&$31.4_{\textcolor{up}{\uparrow 5.0}}$&$27.9_{\textcolor{up}{\uparrow 3.8}}$ \\
 \rowcolor{gray!10} w/ Supervised GRPO&$75.6_{\textcolor{up}{\uparrow 3.0}}$&$28.3_{\textcolor{up}{\uparrow 11.4}}$&$37.8_{\textcolor{up}{\uparrow 4.0}}$&$20.0_{\textcolor{up}{\uparrow 3.3}}$&$55.0_{\textcolor{up}{\uparrow 2.5}}$&$43.3_{\textcolor{up}{\uparrow 4.8}}$&$24.3_{\textcolor{up}{\uparrow 5.7}}$&$23.1_{\textcolor{up}{\downarrow 1.5}}$&$32.9_{\textcolor{up}{\uparrow 6.5}}$&$27.4_{\textcolor{up}{\uparrow 3.3}}$ \\
 w/ Unsupervised EMPO&$74.8_{\textcolor{up}{\uparrow 2.2}}$&$35.3_{\textcolor{up}{\uparrow 18.4}}$&$37.3_{\textcolor{up}{\uparrow 3.5}}$&$16.7_{\textcolor{up}{\uparrow 0.0}}$&$57.5_{\textcolor{up}{\uparrow 5.0}}$&$44.3_{\textcolor{up}{\uparrow 5.8}}$&$25.7_{\textcolor{up}{\uparrow 7.1}}$&$23.1_{\textcolor{up}{\downarrow 1.5}}$&$30.0_{\textcolor{up}{\uparrow 3.6}}$&$26.5_{\textcolor{up}{\uparrow 2.4}}$ \\
\rowcolor{gray!10} w/ \method{} Iter1 &$\textbf{79.0}_{\textcolor{up}{\uparrow 6.4}}$&$38.6_{\textcolor{up}{\uparrow 21.7}}$&$\textbf{42.1}_{\textcolor{up}{\uparrow 8.3}}$&$16.7_{\textcolor{up}{\uparrow 0.0}}$&$50.0_{\textcolor{up}{\downarrow 2.5}}$&$45.3_{\textcolor{up}{\uparrow 6.8}}$&$28.6_{\textcolor{up}{\uparrow 10.0}}$&$23.1_{\textcolor{up}{\downarrow 1.5}}$&$27.9_{\textcolor{up}{\uparrow 1.5}}$&$26.2_{\textcolor{up}{\uparrow 2.1}}$\\
  w/ \method{} Iter2 &$78.4_{\textcolor{up}{\uparrow 5.8}}$&$\textbf{43.8}_{\textcolor{up}{\uparrow 26.9}}$&$39.9_{\textcolor{up}{\uparrow 6.1}}$&$13.3_{\textcolor{up}{\downarrow 3.4}}$&$57.5_{\textcolor{up}{\uparrow 5.0}}$&$46.6_{\textcolor{up}{\uparrow 8.1}}$&$28.6_{\textcolor{up}{\uparrow 10.0}}$&$20.8_{\textcolor{up}{\downarrow 3.8}}$&$37.1_{\textcolor{up}{\uparrow 10.7}}$&$29.1_{\textcolor{up}{\uparrow 5.0}}$\\
\rowcolor{gray!10} w/ \method{} Iter3&$\textbf{79.0}_{\textcolor{up}{\uparrow 6.4}}$&$43.4_{\textcolor{up}{\uparrow 26.5}}$&$41.2_{\textcolor{up}{\uparrow 7.4}}$&$\textbf{26.7}_{\textcolor{up}{\uparrow 10.0}}$&$\textbf{62.5}_{\textcolor{up}{\uparrow 10.0}}$&$\textbf{50.6}_{\textcolor{up}{\uparrow 12.1}}$&$\textbf{31.4}_{\textcolor{up}{\uparrow 12.8}}$&$\textbf{24.6}_{\textcolor{up}{\uparrow 0.0}}$&$\textbf{35.7}_{\textcolor{up}{\uparrow 9.3}}$&$\textbf{30.6}_{\textcolor{up}{\uparrow 6.5}}$\\ \hline
\multicolumn{11}{c}{\textit{Llama-3.2-3B-Instruct}} \\ \hline
\rowcolor{gray!10} Vanilla Base&$47.4$&$21.0$&$13.0$&$3.3$&$22.5$&$21.4$&$27.1$&$17.7$&$15.0$&$18.5$ \\
  w/ Supervised GRPO&$47.2_{\textcolor{up}{\downarrow 0.2}}$&$18.8_{\textcolor{up}{\downarrow 2.2}}$&$14.4_{\textcolor{up}{\uparrow 1.4}}$&$10.0_{\textcolor{up}{\uparrow 6.7}}$&$\textbf{27.5}_{\textcolor{up}{\uparrow 5.0}}$&$23.6_{\textcolor{up}{\uparrow 2.2}}$&$34.3_{\textcolor{up}{\uparrow 7.2}}$&$16.2_{\textcolor{up}{\downarrow 1.5}}$&$19.3_{\textcolor{up}{\uparrow 4.3}}$&$23.3_{\textcolor{up}{\uparrow 4.8}}$ \\ 
\rowcolor{gray!10} w/ Unsupervised EMPO&$47.6_{\textcolor{up}{\uparrow 0.2}}$&$20.6_{\textcolor{up}{\downarrow 0.4}}$&$13.5_{\textcolor{up}{\uparrow 0.5}}$&$6.7_{\textcolor{up}{\uparrow 3.4}}$&$\textbf{27.5}_{\textcolor{up}{\uparrow 5.0}}$&$23.2_{\textcolor{up}{\uparrow 1.8}}$&$30.0_{\textcolor{up}{\uparrow 2.9}}$&$17.7_{\textcolor{up}{\uparrow 0.0}}$&$13.6_{\textcolor{up}{\downarrow 1.4}}$&$18.5_{\textcolor{up}{\uparrow 0.0}}$ \\
 w/ \method{} Iter1 &$47.2_{\textcolor{up}{\downarrow 0.2}}$&$\textbf{22.1}_{\textcolor{up}{\uparrow 1.1}}$&$15.1_{\textcolor{up}{\uparrow 2.1}}$&$10.0_{\textcolor{up}{\uparrow 6.7}}$&$\textbf{27.5}_{\textcolor{up}{\uparrow 5.0}}$&$24.4_{\textcolor{up}{\uparrow 3.0}}$&$32.9_{\textcolor{up}{\uparrow 5.8}}$&$16.9_{\textcolor{up}{\downarrow 0.8}}$&$\textbf{23.6}_{\textcolor{up}{\uparrow 8.6}}$&$24.5_{\textcolor{up}{\uparrow 6.0}}$\\
 \rowcolor{gray!10} w/ \method{} Iter2 &$\textbf{49.0}_{\textcolor{up}{\uparrow 1.6}}$&$\textbf{22.1}_{\textcolor{up}{\uparrow 1.1}}$&$\textbf{16.0}_{\textcolor{up}{\uparrow 3.0}}$&$10.0_{\textcolor{up}{\uparrow 6.7}}$&$22.5_{\textcolor{up}{\uparrow 0.0}}$&$23.9_{\textcolor{up}{\uparrow 2.5}}$&$34.3_{\textcolor{up}{\uparrow 7.2}}$&$21.5_{\textcolor{up}{\uparrow 3.8}}$&$19.3_{\textcolor{up}{\uparrow 4.3}}$&$25.0_{\textcolor{up}{\uparrow 6.5}}$\\

 w/ \method{} Iter3&$47.4_{\textcolor{up}{\uparrow 0.0}}$&$21.7_{\textcolor{up}{\uparrow 0.7}}$&$14.8_{\textcolor{up}{\uparrow 1.8}}$&$\textbf{13.3}_{\textcolor{up}{\uparrow 10.0}}$&$\textbf{27.5}_{\textcolor{up}{\uparrow 5.0}}$&$\textbf{24.9}_{\textcolor{up}{\uparrow 3.5}}$&$\textbf{35.7}_{\textcolor{up}{\uparrow 8.6}}$&$\textbf{23.8}_{\textcolor{up}{\uparrow 6.1}}$&$22.1_{\textcolor{up}{\uparrow 7.1}}$&$\textbf{27.2}_{\textcolor{up}{\uparrow 8.7}}$\\
\Xhline{1.2pt}
\end{tabular}%
}
\caption{Performance comparison of different reinforcement learning strategies on LLMs for mathematical and scientific reasoning tasks. The \textbf{boldfaced} scores represent the \textbf{best} results. The arrows denote the performance change of each method relative to Vanilla Base.}
\label{tab:main_results}
\end{center}
\vspace{-5mm}
\end{table*}

\noindent \textbf{Datasets}. For training, we construct a final dataset consisting of 20,000 samples from DeepMath-103K~\cite{he2025deepmath103klargescalechallengingdecontaminated} with each problem assigned a difficulty level. We randomly select 2,000 samples with difficulty levels ranging from 3.0 to 4.5 as the labeled data, and 18,000 samples with difficulty levels between 5.0 and 10.0 as the unlabeled data to demonstrate the self-evolving process. The selected samples across different difficulty levels are evenly distributed as much as possible. For evaluation, we adopt several well-known mathematical and scientific reasoning benchmarks, including MATH~\citep{NEURIPSDATASETSANDBENCHMARKS2021_be83ab3e}, Minerva MATH~\citep{NEURIPS2022_18abbeef}, OlympiadBench~\citep{he-etal-2024-olympiadbench}, AIME24\footnote{\url{https://huggingface.co/datasets/HuggingFaceH4/aime_2024}}, AMC23\footnote{\url{https://huggingface.co/datasets/AI-MO/aimo-validation-amc}} and GPQA~\citep{rein2023gpqagraduatelevelgoogleproofqa}.

\begin{figure}[t]
    \centering
    \includegraphics[width=\columnwidth]{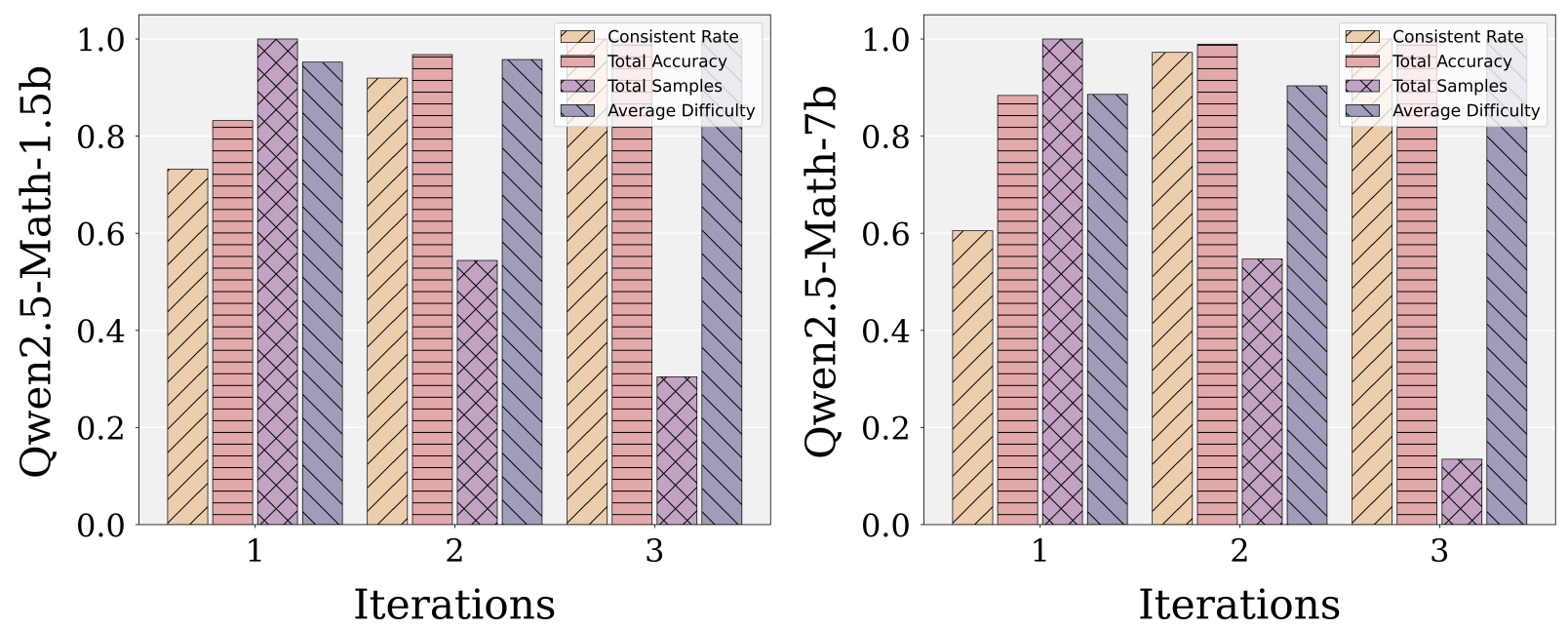}
    \caption{Quality evolution of pseudo-labeled data across three rounds, measured by consistency rate, total accuracy, samples and average difficulty.}
    \label{fig:psedolabel}
    \vspace{-5mm}
\end{figure}

\noindent \textbf{Baseline}. We compare our proposed \method{} to three types of baselines as follows:
\begin{itemize}[left=0pt]
    \vspace{-3mm}
    \item \textbf{Vanilla}: We utilize Qwen2.5-Math-1.5B and Qwen2.5-Math-7B~\citep{yang2024qwen25mathtechnicalreportmathematical} to examine the scalability of \method{}. To assess its generality across architectures, we also include Llama-3.2-3B-Instruct~\citep{grattafiori2024llama3herdmodels}. 
    \vspace{-3mm}
    \item \textbf{Supervised:} We implement supervised GRPO using \texttt{verl}~\citep{Sheng_2025} on labeled data. For every prompt, we sample $8$ responses. The reward is $1$ for a correct answer, $0$ for an incorrect one, and $-0.5$ if no extractable answer is found. 
    \vspace{-3mm}
      \item \textbf{Unsupervised:} We include EMPO~\citep{zhang2025right} as a representative label-free RL baseline. The reward is based on semantic entropy.
\end{itemize}

\begin{figure}[t]
    \centering
    \includegraphics[width=\columnwidth]{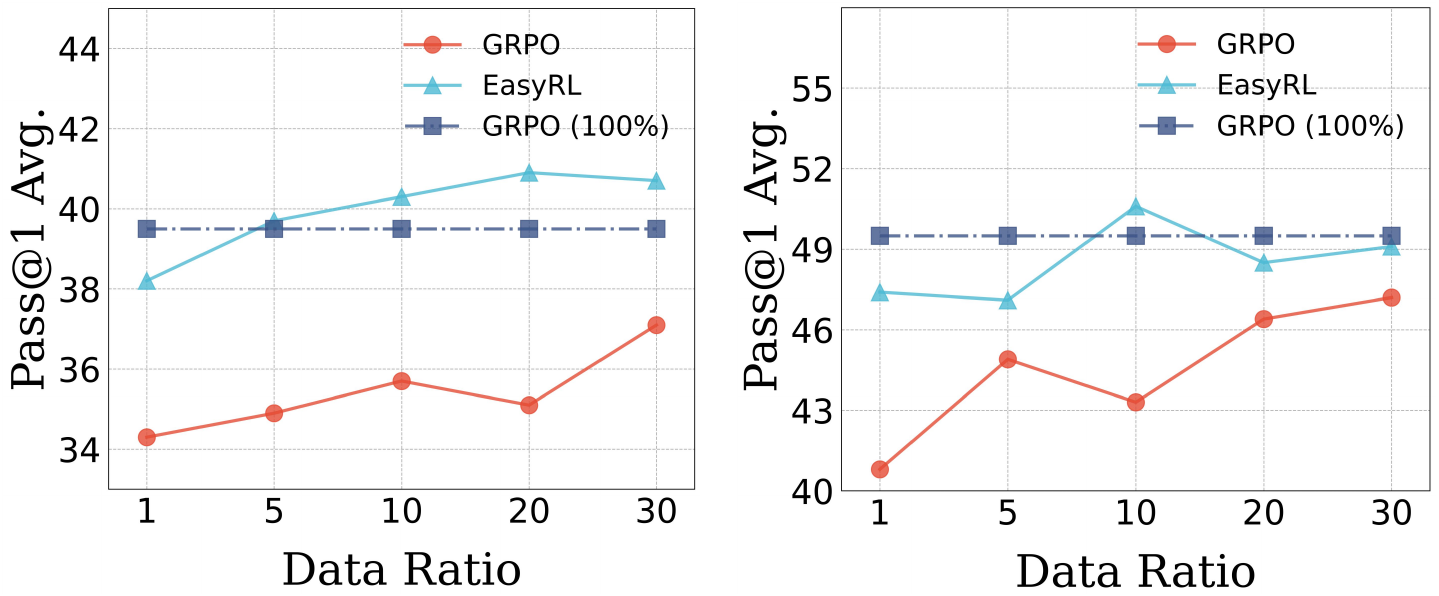}
    \caption{Performance comparison under different proportions of labeled data on Qwen2.5-Math-1.5B (left) and Qwen2.5-Math-7B (right).}
    \label{fig:dataratio}
    \vspace{-5mm}
\end{figure}

\begin{table*}[t]
\renewcommand\arraystretch{1.2}
\label{ablation}
\begin{center}
\resizebox{ \textwidth}{!}{
\begin{tabular}{clccccccc}
\Xhline{1.2pt}
\rowcolor{tabletitle}   \textbf{Model} & \textbf{Variant}           & \textbf{MATH} & \textbf{Minerva} & \textbf{Olympiad Bench}    & \textbf{AIME24} & \textbf{AMC23} & \textbf{GPQA}  & \textbf{Avg.} \\ \Xhline{1.2pt}
& \method{} w/o KT         & $68.2_{\textcolor{down}{\downarrow 3.0}}$    & $29.0_{\textcolor{down}{\downarrow 1.9}}$    & $31.9_{\textcolor{down}{\uparrow 0.6}}$  & $10.0_{\textcolor{down}{\downarrow 3.3}}$   & $52.5_{\textcolor{down}{\downarrow 2.5}}$    & $17.9_{\textcolor{down}{\downarrow 1.5}}$  &  $34.9_{\textcolor{down}{\downarrow 2.0}}$  \\
\rowcolor{gray!10} &  \method{} w/o DC        & $67.6_{\textcolor{down}{\downarrow 3.6}}$    & $25.4_{\textcolor{down}{\downarrow 5.5}}$    & $30.2_{\textcolor{down}{\downarrow 1.1}}$    &  $3.3_{\textcolor{down}{\downarrow 10.0}}$ & $45.0_{\textcolor{down}{\downarrow 10.0}}$    & $18.2_{\textcolor{down}{\downarrow 1.2}}$ & $31.6_{\textcolor{down}{\downarrow 5.3}}$   \\
& \method{} w/o DST         & $69.8_{\textcolor{down}{\downarrow 1.4}}$     & $30.1_{\textcolor{down}{\downarrow 0.8}}$     & $\textbf{32.4}_{\textcolor{down}{\downarrow 1.1}}$   & $3.3_{\textcolor{down}{\downarrow 10.0}}$   & $\textbf{55.0}_{\textcolor{down}{\downarrow 0.0}}$     & $17.6_{\textcolor{down}{\downarrow 1.8}}$ &  $34.7_{\textcolor{down}{\downarrow 2.2}}$   \\ 
\rowcolor{gray!10} \multirow{-4}{*}{Qwen2.5-Math-1.5B}  &   \method{}      & $\textbf{71.2}$    & $\textbf{30.9}$    & $31.3$    & $\textbf{13.3}$ & $\textbf{55.0}$    & $\textbf{19.4}$  & $\textbf{36.9}$ 
  \\ \hline
 & \method{} w/o KT         & $76.2_{\textcolor{down}{\downarrow 2.8}}$    &   $39.3_{\textcolor{down}{\downarrow 4.1}}$  &  $40.9_{\textcolor{down}{\downarrow 0.3}}$ & $20.0_{\textcolor{down}{\downarrow 6.7}}$  &  $57.5_{\textcolor{down}{\downarrow 5.0}}$   &  $28.5_{\textcolor{down}{\downarrow 2.1}}$ & $43.7_{\textcolor{down}{\downarrow 3.5}}$    \\
\rowcolor{gray!10} &  \method{} w/o DC        &  $76.4_{\textcolor{down}{\downarrow 2.6}}$   & $39.0_{\textcolor{down}{\downarrow 4.4}}$   &  $38.1_{\textcolor{down}{\downarrow 3.1}}$    & $23.3_{\textcolor{down}{\downarrow 3.4}}$  & $57.5_{\textcolor{down}{\downarrow 5.0}}$  &  $25.3_{\textcolor{down}{\downarrow 5.3}}$ & $43.3_{\textcolor{down}{\downarrow 3.9}}$\\
& \method{} w/o DST         & $78.4_{\textcolor{down}{\downarrow 0.6}}$  &   $\textbf{44.1}_{\textcolor{down}{\uparrow 0.7}}$  & $37.8_{\textcolor{down}{\downarrow 3.4}}$  & $\textbf{26.7}_{\textcolor{down}{\downarrow 0.0}}$  & $57.5_{\textcolor{down}{\downarrow 5.0}}$    & $29.1_{\textcolor{down}{\downarrow 1.5}}$ &  $45.6_{\textcolor{down}{\downarrow 1.6}}$   \\ 
\rowcolor{gray!10} \multirow{-4}{*}{Qwen2.5-Math-7B} &  \method{}      & $\textbf{79.0}$   &   $43.4$  &  $\textbf{41.2}$   & $\textbf{26.7}$ & $\textbf{62.5}$    & $\textbf{30.6}$ & $\textbf{47.2}$
    \\
\Xhline{1.2pt}
\end{tabular}%
}
\caption{Ablation study via performance comparison of different variants on \method{}. The \textbf{boldfaced} scores represent the \textbf{best} results. The arrows denote the performance change of each variant relative to \method{}.}
\label{tab:ablation_study}
\end{center}
\vspace{-5mm}
\end{table*}

\vspace{-2mm}
\noindent \textbf{Implementation Details.} For training, we implement GRPO via~\citep{Sheng_2025} and run experiments on $8\times140\ \mathrm{GB}$ H200 GPUs. We perform two inferences per sample for consistency-based selection. For evaluation, we use zero-shot prompting with greedy decoding and report \texttt{pass@1} accuracy. 



\subsection{Main Results}

\noindent \textbf{Performance on Mathematical Reasoning Tasks.}
We report the main results of \method{} in Table~\ref{tab:main_results} and summarize several observations. Firstly, \method{} consistently outperforms both supervised and unsupervised RL baselines across all backbone models. Notably, with only 10\% of easy labeled data, it achieves the largest improvements of 7.7\% on Qwen2.5-Math-1.5B, 12.1\% on Qwen2.5-Math-7B, and 3.5\% on LLaMA3.2-3B-Instruct. Secondly, the iterative self-evolution process leads to a steady improvement in model performance, e.g., Qwen2.5-Math-7B improves by 5.3\% from Iter1 to Iter3. Thirdly, \method{} demonstrates strong transferability, effectively enhancing the reasoning capability of language models across different architectures and model scales.


\noindent \textbf{Performance on Scientific Reasoning Tasks.} Previous studies have shown that training language models on reasoning-intensive domains such as mathematics can enhance their general reasoning capabilities across other fields~\citep{huan2025doesmathreasoningimprove}. To evaluate the out-of-domain (OOD) generalization of our method, we train \method{} on math datasets and assess its transfer performance on biology, chemistry, and physics benchmarks. As shown in Table~\ref{tab:main_results}, \method{} consistently improves over the base model, achieving an average gain of 17.9\% on Qwen2.5-Math-1.5B, 6.5\% on Qwen2.5-Math-7B, and 8.7\% on LLaMA3.2-3B-Instruct. These results demonstrate that \method{} has strong generalization and exhibits OOD reasoning capability.

\begin{tcolorbox}[
    enhanced, 
    colback=white, 
    colframe=deepgreen, 
    boxrule=1pt, 
    arc=5pt, 
    left=5pt, right=5pt, 
    top=5pt, bottom=5pt, 
    width=\linewidth 
]
\textbf{\textcolor{deepgreen}{Finding 1}}: \method{} achieves robust performance gains: it outperforms supervised and unsupervised RL baselines, improves steadily through iterative self-evolution, and transfers effectively to out-of-domain scientific tasks.

\end{tcolorbox}

\begin{figure*}[t]
    \centering
    \includegraphics[width=\linewidth]{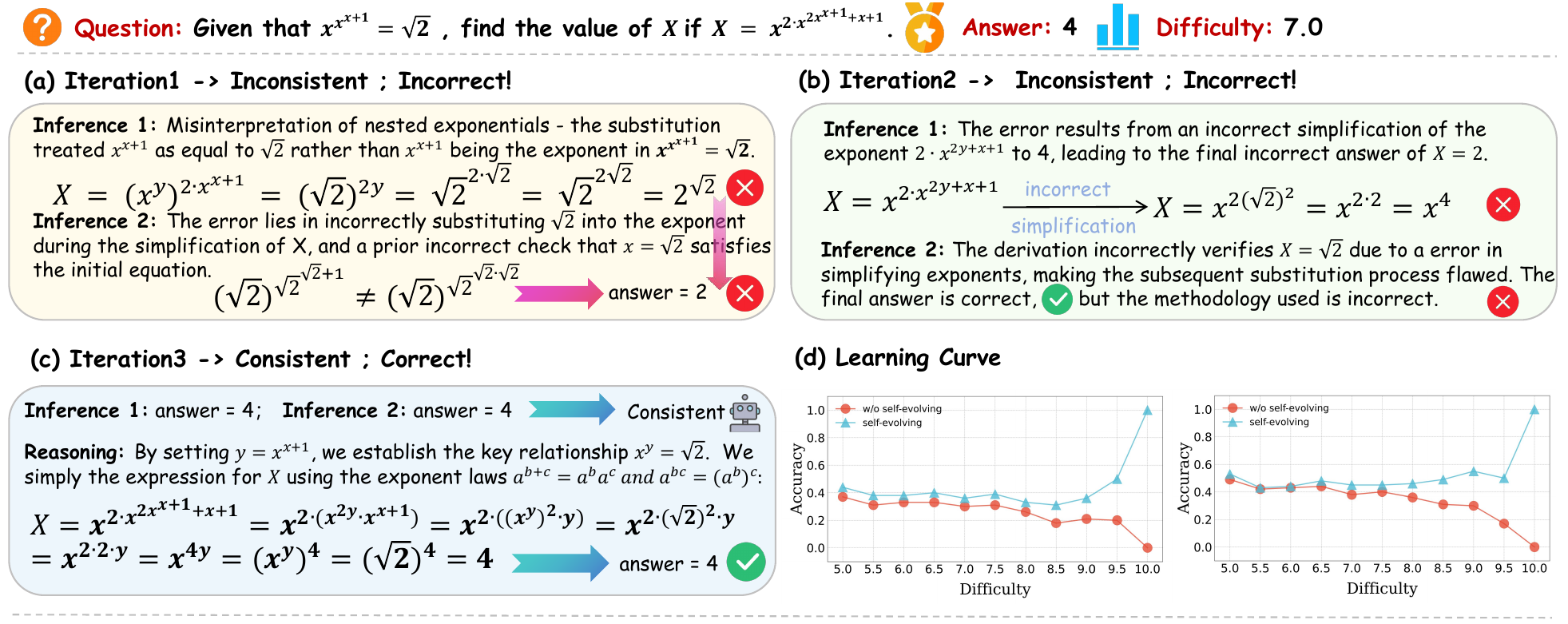}
    \caption{Case Study. Comparison of the pseudo-labeling process, generated by: (a) \method{} Iteration 1, (b) \method{} Iteration 2, and (c) \method{} Iteration 3. Panel (d) shows the learning curve.}
    \vspace{-7mm}
    \label{fig:case}
\end{figure*}

\vspace{-5mm}
\subsection{Further Analysis}

\noindent \textbf{Pseudo-label Quality Analysis.} To assess the quality of pseudo-labels across three iterations, we adopt four metrics: Consistent Rate, the proportion of samples that yield identical outputs across multiple reasoning attempts; Total Accuracy, the fraction of correctly pseudo-labeled samples; Total Samples, the number of selected pseudo-labeled samples in each round; and Average Difficulty, the mean difficulty of samples in \(\mathcal{D}_{\text{unlabel\_selected}}^{(i)}\). Figure~\ref{fig:psedolabel} shows the normalized trends of Consistent Rate and Total Accuracy over \(\mathcal{D}_{\text{unlabel}}\), as well as Total Samples and Average Difficulty in \(\mathcal{D}_{\text{unlabel\_selected}}^{(i)}\), for Qwen2.5-Math-1.5B and 7B. 

From these results, we draw three observations: \ding{182} Across iterations, both Consistent Rate and Total Accuracy on \(\mathcal{D}_{\text{unlabel}}\) steadily increase, indicating that models improve through self-evolution. Consistent Rate and Total Accuracy are positively correlated, which aligns with prior studies~\citep{englesson2021consistencyregularizationimproverobustness,10.5555/3495724.3495775}; \ding{183} Total Samples decreases over iterations, as each round operates on the remaining subset from the previous iteration; and \ding{184} Average Difficulty increases, with the overall average difficulty on \(\mathcal{D}_{\text{unlabel}}\) being 6.9274, while in iteration 3 it reaches 7.0282 for the 1.5B model and 7.4256 for the 7B model. This pattern suggests that the model’s learning process gradually shifts from easier to more difficult cases.

\begin{tcolorbox}[
    enhanced, 
    colback=white, 
    colframe=deepblue, 
    boxrule=1pt, 
    arc=5pt, 
    left=5pt, right=5pt, 
    top=5pt, bottom=5pt, 
    width=\linewidth 
]
\textbf{\textcolor{deepblue}{Finding 2}}: \method{} exhibits a self-evolving pseudo-labeling process, with steadily improving pseudo-label quality and increasing focus on more difficult samples.
\end{tcolorbox}

\noindent \textbf{Data Proportion Analysis.} We evaluate the performance of supervised GRPO and \method{} using 1\%, 5\%, 10\%, 20\%, and 30\% of \(\mathcal{D}_{\text{label}}/\mathcal{D}\), as well as supervised GRPO trained on 100\% labeled data, on Qwen2.5-Math-1.5B and Qwen2.5-Math-7B (see Figure~\ref{fig:dataratio}). Performance is measured as average accuracy across five math reasoning datasets. We draw two conclusions. First, \method{} consistently outperforms supervised GRPO when trained with the same amount of labeled data. Second, using only 10\% of easy labeled data, \method{} already surpasses the performance of supervised GRPO trained on 100\% labeled data, achieving 40.3 compared to 39.5 on Qwen2.5-Math-1.5B and 50.6 compared to 49.5 on Qwen2.5-Math-7B. Additional results are provided in Appendix~\ref{sec:further_analysis}.

\begin{tcolorbox}[
    enhanced, 
    colback=white, 
    colframe=purple, 
    boxrule=1pt, 
    arc=5pt, 
    left=5pt, right=5pt, 
    top=5pt, bottom=5pt, 
    width=\linewidth 
]
\textbf{\textcolor{purple}{Finding 3}}: \method{} demonstrates strong data efficiency: with only 10\% easy labeled data, it surpasses supervised GRPO trained on 100\% labeled data and consistently outperforms GRPO under equal labeled data budgets.
\end{tcolorbox}

\noindent \textbf{Ablation Study.} To validate the contribution of each core component, we conduct ablation studies on Qwen2.5-Math-1.5B and Qwen2.5-Math-7B, as shown in Table~\ref{tab:ablation_study}. We compare three variants: \ding{182} \method{} w/o Knowledge Transfer (KT): a variant that removes the supervised GRPO warm-up on labeled data; \ding{183} \method{} w/o Divide and Conquer (DC): a variant that applies unsupervised RL training directly after $\pi_{\text{warm}}$; and \ding{184} \method{} w/o Difficulty-Progressive Self-Training (DST): a variant that performs only one round of pseudo-labeling and selection after $\pi_{\text{warm}}$, jointly training on $\mathcal{D}_{\text{label}} \cup \mathcal{D}_{\text{unlabel\_selected}}^{(1)}$ to obtain the final model $\pi_{\text{final}}$. To ensure fair comparison, we select the top 70\% pseudo-labeled samples with the lowest dynamic entropy for training. The results show that removing any of the Knowledge Transfer, Divide and Conquer, or Difficulty-Progressive Self-Training components leads to an average performance drop of 2.8\%, 4.60\%, and 1.9\% on two models, demonstrating that each component plays an indispensable role in the overall performance of \method{}.


\noindent \textbf{Case Study.}
We present a case study on a problem from $\mathcal{D}_{\text{unlabel}}$ (see Figure~\ref{fig:case}). Across iterations, the model shows a refinement process: Iter~1 yields an inconsistent and incorrect solution; in Iter~2, one inferred answer is correct but the reasoning process remains flawed; and Iter~3 finally produces a consistent and correct answer. We further report learning curves for Qwen2.5-Math-1.5B (left) and Qwen2.5-Math-7B (right), showing accuracy on $\mathcal{D}_{\text{unlabel}}$ after pseudo-labeling and selection across difficulty levels. While the non–self-evolving baseline exhibits a steep accuracy drop as difficulty increases, the self-evolving model maintains more stable performance and achieves gains in the 9.0–10.0 difficulty range, highlighting the effectiveness of \method{} under challenging reasoning conditions.
\section{Conclusion}

In this work, we present \method{}, a data-efficient reinforcement learning framework that enables LLMs to self-evolve from limited easy labeled samples. \method{} first initializes a reliable policy via supervised reinforcement learning, then constructs high-quality pseudo-labeled data through a divide-and-conquer strategy combining consistency and reflection. Finally, \method{} progressively refines the model by incorporating difficult samples in a staged training loop. Experiments show that \method{} outperforms strong baselines and significantly improves pseudo-label quality, enabling self-evolving reasoning in LLMs.

\section{Limitations}
Despite its promising performance, \method{} still faces several limitations. Our current experiments mainly focus on domains with verifiable rewards, such as mathematical reasoning. Extending \method{} to open-ended tasks with non-verifiable or subjective rewards, including creative writing and scientific research, remains a challenging but valuable future direction. Moreover, future work may apply \method{} to real-world scenarios such as embodied intelligence or social simulation.

\section{Ethical Statement}
This work adheres to ethical research and deployment principles. All experiments are conducted on publicly available datasets, and no private, sensitive, or personally identifiable data are used. Our \method{} aims to improve model reasoning ability in a transparent and reproducible manner.

Nevertheless, as with all self-evolving AI systems, care should be taken to prevent potential misuse or unintended reinforcement of biased behaviors. We encourage future research to integrate fairness-aware reward functions and continuous human oversight to ensure that self-evolving language models remain aligned with human values and societal norms.

\section{Acknowledgments}
The work of Zhiyin Yu, Bo Zhang, and Lei Bai was supported by the Shanghai Artificial Intelligence Laboratory. The authors thank the anonymous reviewers for their valuable comments and suggestions.

\bibliography{custom}

\appendix

\section{Experimental Settings}
\label{sec:exp_settings}

\subsection{Parameter Settings}

Following previous work, we build on the SimpleRL codebase~\citep{zeng2025simplerl} and adopt the standard evaluation prompt~\citep{zhang2025right}: ``Let’s think step by step and output the final answer within \verb|\boxed{}|.'' During inference, we use a greedy decoding configuration with temperature set to 0, top-$p$ set to 1, and a fixed random seed of 0. All experiments are conducted with the vLLM framework on 2 x NVIDIA H200 GPUs (140GB).

For RL training, we adopt the GRPO algorithm with a maximum sequence length of 4096. The training batch size is set to 8, with a mini-batch size of 2 and a micro-batch size of 1. The actor and critic are optimized using learning rates of $5\times10^{-7}$ and $9\times10^{-6}$, respectively. All experiments are conducted on eight NVIDIA H200 GPUs (140GB each), and the model is trained for a single epoch.

\subsection{Instruction Settings}


\newtcolorbox{math_template}{
    colback=blue!10,    
    colframe=blue!50,      
    title=Mathematical Reasoning Training and Evaluation Template,       
    fonttitle=\bfseries\color{white},
    boxrule=1pt,        
    arc=3pt,            
    left=6pt,           
    right=6pt,          
    top=4pt,            
    bottom=4pt,         
    before upper={\parindent0pt}, 
    coltitle=white,      
}

\begin{math_template}

<|im\_start|> system

Please reason step by step, and output your final answer within \textbackslash boxed\{\}.

<|im\_end|>

<|im\_start|>user

\{Question\} Let's think step by step and output the final answer within \textbackslash boxed\{\}.

<|im\_end|>

<|im\_start|>assistant
\end{math_template}

\vspace{5mm}

\newtcolorbox{grqa_test_template}[1][]{  
    colback=orange!10,
    colframe=orange!50,
    title={#1},  
    fonttitle=\bfseries\color{white},
    boxrule=1pt,
    arc=3pt,
    left=6pt,
    right=6pt,
    top=4pt,
    bottom=4pt,
    before upper={\parindent0pt},
    coltitle=white
}

\begin{grqa_test_template}[GPQA Test Prompt]
<|im\_start|>system

Reason step by step, and output your final answer (A, B, C, or D) within \textbackslash boxed\{\}.

<|im\_end|>

<|im\_start|>user

\{Question\} Reason step by step and output the final answer (A, B, C, or D) within \textbackslash boxed\{\}.

<|im\_end|>

<|im\_start|>assistant
\end{grqa_test_template}

\vspace{3mm}

\newtcolorbox{natural_train_template}{
    colback=pink!10,    
    colframe=pink,      
    title=Reflection-based Resolution,       
    fonttitle=\bfseries\color{white},
    boxrule=1pt,        
    arc=3pt,            
    left=6pt,           
    right=6pt,          
    top=4pt,            
    bottom=4pt,         
    before upper={\parindent0pt}, 
    coltitle=white,      
}
\begin{natural_train_template}
You are given multiple proposed answers to a math problem.  
Your task is to carefully examine these answers and determine whether any of them is correct.

- If one of the proposed answers is correct, return it as the final answer.

- If none of the proposed answers is correct, re-solve the problem step-by-step and provide the correct answer.

- Always show the final answer clearly inside \textbackslash boxed\{\}.

Question:
\{question\}

Proposed Answers:
\{answers\}

Now, please reflect on the answers above and give the final correct answer in \textbackslash boxed\{\}.
\end{natural_train_template}

\vspace{2mm}
\section{Further Analysis}
\label{sec:further_analysis}

Table~\ref{tab:num_infers} reports the results of the consistency-based selection during pseudo-labeling, where we vary the number of inferences (N) as 2, 3, and 4. We compare the consistent rate, average entropy, total number of selected samples, and accuracy. As expected, a larger N leads to a lower consistent rate. As discussed in Section~3.3, the consistent rate is positively correlated with the total accuracy. Therefore, we set N=2 as the default configuration, which achieves a good balance between performance and computational efficiency.

\vspace{2mm}

\begin{table}[h]
\centering
\resizebox{0.95 \columnwidth}{!}{
\begin{tabular}{rcccc}
\Xhline{1.2pt}
\rowcolor{tabletitle}  N & Cons Rate & Entropy & Samples & Accuracy  \\ \Xhline{1.2pt}
2 & 0.1653 & 0.1550 & 7483 &  0.3186 \\
\rowcolor{gray!10} 3 & 0.0983 & 0.1593 &  6638 & 0.3039   \\
 4 & 0.0706 & 0.1613 & 6289 & 0.2822 \\
\Xhline{1.2pt}
\end{tabular}
}
\caption{Analysis of the number of inferences (N) in consistency-based selection.}
\label{tab:num_infers}
\end{table}

We provide analysis on the threshold setting (0.1–0.5) using Qwen2.5-Math-1.5B across multiple benchmarks. From the results shown in Table~\ref{tab:threshold}, we make the following observations: (1) \method{} demonstrates robust performance across various settings, indicating low sensitivity to the threshold. (2) The threshold determines the percentage of inconsistent samples with the lowest entropy that are selected in each iteration. As the threshold increases, more samples are selected. However, a higher threshold also introduces samples with greater uncertainty or noise. Therefore, when the threshold becomes too large, performance drops, suggesting that the excessive inclusion of uncertain samples negatively affects training. (3) As the threshold decreases, only a limited number of samples are selected per iteration. This restricts the model’s exposure to diverse training signals and leads to suboptimal improvement. We choose 0.3 as the default value, as it provides the best balance between sample quality and quantity.

\begin{table}[h]
\centering
\resizebox{ \columnwidth}{!}{
\begin{tabular}{rcccccc}
\Xhline{1.2pt}
\rowcolor{tabletitle}  Thres & MATH & Minerva & Olympiad & AIME24 & AMC23& Avg.  \\ \Xhline{1.2pt}
0.1 & 69.8 & 26.5 & 30.7 &  13.3 &52.5 &38.6 \\
\rowcolor{gray!10} 0.2 & 70.0 & 30.1 &  34.2 & 13.3 &52.5&40.0   \\
 0.3 & 71.2 & 30.9 & 31.3 & 13.3&55.0&40.3 \\
 0.4&71.0&28.3&32.7&6.7&50.0&37.7\\
 0.5&68.2&31.6&31.6&6.7&50.0&37.6\\
\Xhline{1.2pt}
\end{tabular}
}
\caption{Sensitivity analysis of the threshold.}
\label{tab:threshold}
\end{table} 

We also extend our experiments to five self-training iterations using Qwen2.5-Math-1.5B. The results are summarized in Table~\ref{tab:iteration}. From Iter 1 to Iter 5, we observe the following: (1) Performance steadily improves from Iteration 1 to Iteration 3, demonstrating that iterative self-training effectively enhances the model’s reasoning capability in the early stages. (2) After Iteration 3, performance gradually converges, with the average score showing only marginal improvement or slight decline. This suggests that excessive self-training may introduce accumulated noise from pseudo-labels, leading to minor performance degradation. 

\begin{table}[h]
\centering
\resizebox{ \columnwidth}{!}{
\begin{tabular}{rcccccc}
\Xhline{1.2pt}
\rowcolor{tabletitle}  Iter & MATH & Minerva & Olympiad & AIME24 & AMC23& Avg.  \\ \Xhline{1.2pt}
1 & 69.8 & 28.7 & 31.0 &  3.3 &52.5 &37.1 \\
\rowcolor{gray!10} 2 & 70.0 & 31.6 &  31.3 & 10.0 &50.0&38.6   \\
3 & 71.2 & 30.9 & 31.3 & 13.3&55.0&40.3 \\
4&73.8&32.7&33.0&10.0&52.5&40.4\\
5&74.2&32.0&33.8&10.0&47.5&39.5\\
\Xhline{1.2pt}
\end{tabular}
}
\caption{Sensitivity analysis of iterations.}
\label{tab:iteration}
\end{table}

Table~\ref{tab:all_data_ratio} presents the sensitivity analysis of our method under different ratios of labeled data. The GRPO variants represent standard RL trained with varying amounts of labeled samples, while the \method{} variants apply our self-evolving mechanism under the same limited supervision. Notably, \method{} achieves comparable or even superior performance to fully supervised GRPO, demonstrating the effectiveness of our data-efficient RL even with only 1--10\% labeled data.

Table~\ref{tab:pse} summarizes the self-evolution dynamics across three pseudo-labeling iterations. As iterations proceed, the consistency rate and accuracy on $D_{\text{unlabel}}$ consistently improve, indicating more reliable pseudo-label generation. Meanwhile, the average difficulty of the selected samples gradually increases, showing that the model progressively shifts its focus from easy to harder problems. In contrast, the accuracy on $D_{\text{selected}}$ declines as samples become more challenging.

\section{Related Work}
\textbf{Curriculum Learning in LLMs.} Curriculum Learning trains models on samples of increasing difficulty to improve convergence and generalization, and has been widely applied in computer vision (CV)~\cite{deng2025boostinggeneralizationreasoningvision}, natural language processing (NLP)~\cite{platanios-etal-2019-competence}, and reinforcement learning (RL)~\cite{wang2025dumpautomateddistributionlevelcurriculum}. Traditional curriculum learning methods typically rely on predefined static difficulty metrics to sort labeled training data offline~\cite{lee-etal-2024-instruction,kimiteam2025kimik15scalingreinforcement}. More recent work explores dynamic or adaptive curriculum learning strategies, where sample difficulty is re-estimated during training to adjust batch ordering~\cite{zhang-etal-2025-learning-like}. In contrast, \method{} targets a data-scarce setting, where models evolve from limited easy samples to more difficult reasoning tasks. For methodology, \method{} transfers knowledge from easy samples and progressively incorporates unlabeled data via a divide-and-conquer strategy, enabling difficulty-progressive self-training to enhance reasoning ability.

\section{Case Study}
As shown in Table~\ref{tc:amc23q13}, the model demonstrates an instance of self-reflection. Initially, it explores potential solutions by testing the symmetric case $a=b$ and then evaluates specific candidate values such as $a=1$ or $b=1$. Upon finding that these attempts yield no valid solutions, the model revisits its reasoning strategy, systematically reconsidering the structure of the equation and leveraging symmetry to identify the correct solution $(a, b) = (\frac{1}{2}, \frac{1}{2})$. 

\onecolumn

\clearpage

\begin{table*}[t]
\renewcommand\arraystretch{1.2}
\label{sensitivity}
\begin{center}
\resizebox{0.95 \textwidth}{!}{
\begin{tabular}{clcccccc}
\Xhline{1.2pt}
\rowcolor{tabletitle}   \textbf{Model} & \textbf{Variant}           & MATH-500 & Minerva MATH & Olympiad Bench    & AIME24 & AMC23 & Avg.   \\ \Xhline{1.2pt}
& 1\% GRPO         & 65.2    & 19.1    & 29.8  & 10.0   & 47.5    & 34.3      \\
\rowcolor{gray!10} &  5\% GRPO        & 66.4    & 24.6    & 28.3    &  10.0 & 45.0    & 34.9   \\
 & 10\% GRPO&66.4&27.9&30.7&3.3&50.0&35.7\\
\rowcolor{gray!10} &  20\% GRPO         & 68.2     & 26.8     & 29.5   & 3.3   & 47.5     & 35.1    \\ 
  &   30\% GRPO     & 69.4    & 27.9    & 31.3    & 6.7 & 50.0    & 37.1    \\ 
  \rowcolor{gray!10} & 100\% GRPO &72.6&32.7&33.9&3.3&55.0&39.5 \\
 &   1\% \method{}  & 69.8    &  31.2   &  33.2 & 6.7  &  50.0   & 38.2     \\
\rowcolor{gray!10} & 5\% \method{}  &  71.0   & 30.9   & 33.3    & 13.3  & 50.0  & 39.7  \\
 &  10\% \method{}&71.2&30.9&31.3&13.3&55.0&40.3\\
\rowcolor{gray!10} &  20\% \method{}   &  71.4 & 30.5   & 32.0  & 13.3  &   57.5  &   40.9  \\ 
 \multirow{-11}{*}{Qwen2.5-Math-1.5B} &  30\% \method{}  &  71.6  &  31.2   &  34.8   & 13.3 &  52.5   & 40.7    \\ \hline
& 1\% GRPO         & 71.0    & 21.7   & 34.7  & 26.7  &  50.0  & 40.8     \\
\rowcolor{gray!10} &  5\% GRPO        & 73.8   &  32.0   & 35.6   & 23.3  &  60.0  & 44.9   \\
 & 10\% GRPO&75.6&28.3&37.8&20.0&55.0&43.3\\
\rowcolor{gray!10} &  20\% GRPO         &  77.8   &  33.5    & 40.6 & 20.0  & 60.0     &  44.9   \\ 
  &  30\% GRPO     &  78.0   &  39.3   & 39.4    & 16.7 &  62.5   &  47.2   \\ 
\rowcolor{gray!10} &  100\% GRPO &80.8&39.3&42.2&20.0&65.0&49.5 \\
 & 1\% \method{}  &  77.4  &   42.3  &  39.1 &  13.3 &  65.0  & 47.4     \\
 \rowcolor{gray!10} &  5\% \method{}  & 77.8    &  41.9  & 40.0    & 13.3  & 62.5  & 47.1  \\
 & 10\% \method{}&79.0&43.4&41.2&26.7&62.5&50.6\\
\rowcolor{gray!10} &  20\% \method{}   & 79.0  &  39.3  & 38.5  & 23.3  &   62.5  &  48.5   \\ 
 \multirow{-11}{*}{Qwen2.5-Math-7B} & 30\% \method{}  & 80.6   &  39.7   & 38.7    & 16.7&70.0     & 49.1   \\ \hline
\Xhline{1.2pt}
\end{tabular}%
}
\caption{Performance comparison of GRPO and \method{} under different labeled data ratios.}
\vspace{-7mm}
\label{tab:all_data_ratio}
\end{center}
\end{table*}

\begin{table*}[t]
\centering
\resizebox{\textwidth}{!}{
\begin{tabular}{lcccccc}
\Xhline{1.2pt}
\rowcolor{tabletitle} Models & Iteration & Data & Consistent Rate & Total Samples & Accuracy & Difficulty  \\ \Xhline{1.2pt}
\rowcolor{gray!10} &  1 & Dunlabel & 0.1653 & 7483 &  0.3186 & 6.6933 \\ 
& &  Dunlabel & 0.2077 & 8017 & 0.3706 & 6.9274 \\
\rowcolor{gray!10} & \multirow{-2}{*}{2}& Dselected1 & 0.1251 & 4078 &0.2469 &6.7308 \\ 
& & Dunlabel & 0.2259 & 8247 &0.3829 &6.9274 \\
\rowcolor{gray!10} \multirow{-5}{*}{Qwen2.5-Math-1.5B}& \multirow{-2}{*}{3}&Dselected2 &0.0777&2284&0.1375&7.0282\\ \hline
&  1 & Dunlabel & 0.1939 & 7844 & 0.4106 & 6.5788 \\ 
\rowcolor{gray!10} &  &  Dunlabel & 0.3117 & 9328 & 0.4597 & 6.5110 \\
&\multirow{-2}{*}{2}& Dselected1 & 0.1747 & 4289 &0.2949 &6.7093 \\ 
\rowcolor{gray!10} &  & Dunlabel & 0.3205 & 9438 &0.4646 &6.5388 \\
\multirow{-5}{*}{Qwen2.5-Math-7B}& \multirow{-2}{*}{3}&Dselected2 &0.0614&1055&0.1412&7.4256\\ \hline
&  1 & Dunlabel & 0.0982 & 6637 & 0.1770 & 6.8001  \\ 
\rowcolor{gray!10} &  &  Dunlabel & 0.2687 & 8785 & 0.2444& 6.7625\\
&\multirow{-2}{*}{2}& Dselected1 & 0.1368 & 4497 & 0.1841& 6.8611  \\ 
\rowcolor{gray!10} &  & Dunlabel & 0.3168 & 9392  & 0.2528  &6.7593\\
\multirow{-5}{*}{Llama-3.2-3B-Instruct}& \multirow{-2}{*}{3}&Dselected2 &0.1486 &2774&0.1388&6.9638\\ 
\Xhline{1.2pt}
\end{tabular}
}
\caption{Self-evolution dynamics during iterative pseudo-labeling.}
\vspace{-3mm}
\label{tab:pse}
\end{table*} 

\clearpage

\newtcolorbox{test_case}{
    colback=gray!10,    
    colframe=gray!50,   
    title=Qwen2.5-Math-7B's Answer to the 13-th Question in AMC23, 
    fonttitle=\bfseries\color{white},
    boxrule=1pt,        
    arc=3pt,            
    left=6pt,           
    right=6pt,          
    top=4pt,            
    bottom=4pt,         
    before upper={\parindent0pt}, 
    coltitle=white,     
    lower separated=true, 
    lowerbox=visible,   
    before lower={\textcolor{gray!70}{\rule{\linewidth}{0.5pt}\vspace{4pt}}}, 
    fontupper=\color{black}, 
    fontlower=\color{black}  
}

\begin{test_case}
\label{tc:amc23q13}
\textbf{Question: How many ordered pairs of positive real numbers $(a,b)$ satisfy the equation $(1+2a)(2+2b)(2a+b) = 32ab$?}

\vspace{0.4pt}

\textbf{Response:} To solve the equation $(1+2a)(2+2b)(2a+b) = 32ab$ for positive real numbers $a$ and $b$, we can use the method of substitution and symmetry. Let's start by expanding and simplifying the left-hand side of the equation.

First, let's expand the left-hand side:
\[
(1+2a)(2+2b)(2a+b) = (1+2a)(2(1+b)(2a+b)) = 2(1+2a)(2a+b)(1+b).
\]
...
Now, let's check if there are any obvious solutions. One way to do this is to check if $a = b$. If $a = b$, then the equation becomes:
\[
(1+2a)(2a+a)(1+a) = 16a^2,
\]
...
Since $a$ is a positive real number, $a \neq 0$. So we need to solve the quadratic equation:
\[
6a^2 - 7a + 3 = 0.
\]
The discriminant of this quadratic equation is:
\[
\Delta = (-7)^2 - 4 \cdot 6 \cdot 3 = 49 - 72 = -23,
\]
which is negative. \textcolor{red}{Therefore, the quadratic equation has no real solutions. This means that $a = b$ does not provide any valid solutions. Next, let's check if there are any other solutions.} We can use the method of symmetry and substitution. Let's try $a = 1$ and solve for $b$:
\[
(1+2 \cdot 1)(2+2b)(2 \cdot 1 + b) = 32 \cdot 1 \cdot b,
\]
...
The discriminant of this quadratic equation is:
\[
\Delta = (-7)^2 - 4 \cdot 3 \cdot 6 = 49 - 72 = -23,
\]
which is negative. Therefore, the quadratic equation has no real solutions. This means that $a = 1$ does not provide any valid solutions.

By symmetry, we can try $b = 1$ and solve for $a$:
\[
(1+2a)(2+2 \cdot 1)(2a+1) = 32a \cdot 1,
\]
...
So the solution is: $a = \frac{1}{2}$.
By symmetry, if $a = \frac{1}{2}$, then $b = \frac{1}{2}$ is also a solution. Therefore, the only solution is $(a, b) = \left(\frac{1}{2}, \frac{1}{2}\right)$.

Thus, there is only one ordered pair of positive real numbers $(a, b)$ that satisfies the equation. The final answer is: \(\boxed{1}\).

\end{test_case}

\end{document}